\documentclass[sigconf,natbib=false,nonacm]{acmart}
\AtBeginDocument{%
  }

\RequirePackage[
  datamodel=acmdatamodel,
  style=acmnumeric,
  ]{biblatex}
\usepackage{algorithm}
\usepackage{algorithmic}
\usepackage{multirow}

\begin{document}

\title{Beyond Known Facts: Generating Unseen Temporal Knowledge to Address Data Contamination in LLM Evaluation}

\author{Arthur Amalvy, Hen-Hsen Huang}
\email{arthuramalvy@as.edu.tw, hhhuang@iis.sinica.edu.tw}
\affiliation{
  \institution{Institute of Information Science, Academia Sinica}
  \city{Taipei}
  \country{Taiwan}
}


\begin{abstract}
The automatic extraction of information is important for populating large web knowledge bases such as Wikidata. 
The temporal version of that task, temporal knowledge graph extraction (TKGE), involves extracting temporally grounded facts from text, represented as semantic quadruples (subject, relation, object, timestamp). Many recent systems take advantage of large language models (LLMs), which are becoming a new cornerstone of the web due to their performance on many tasks across the natural language processing (NLP) field.
Despite the importance of TKGE, existing datasets for training and evaluation remain scarce, and contamination of evaluation data is an unaddressed issue, potentially inflating LLMs' perceived performance due to overlaps between training and evaluation sets.
To mitigate these challenges, we propose a novel synthetic evaluation dataset constructed from predicted future, previously unseen temporal facts, thereby eliminating contamination and enabling robust and unbiased benchmarking.
Our dataset creation involves a two-step approach: (1) Temporal Knowledge Graph Forecasting (TKGF) generates plausible future quadruples, which are subsequently filtered to adhere to the original knowledge base schema; (2) LLMs perform quadruple-to-text generation, creating semantically aligned textual descriptions. 
We benchmark Extract, Define and Canonicalize (EDC), a state-of-the-art LLM-based extraction framework, demonstrating that LLM performance decreases when evaluated on our dataset compared to a dataset of known facts. 
We publicly release our dataset consisting of 4.2K future quadruples and corresponding textual descriptions, along with the generation methodology, enabling continuous creation of unlimited future temporal datasets to serve as long-term, contamination-free benchmarks for TKGE.
\end{abstract}

\ccsdesc[500]{Computing methodologies~Information extraction}
\ccsdesc[500]{Information systems~Web mining}
\ccsdesc[500]{Computing methodologies~Language resources}

\keywords{Temporal Knowledge Graph Extraction, Large Language Models, Knowledge Base}

\received{7 October 2025}

\maketitle

\section{Introduction}
\label{sec:introduction}
Large knowledge bases such as Wikidata~\parencite{Vrandecic2014-wikidata} are a pillar of the web. 
They provide structured, machine-readable data, which can power applications, search engines or AI training and inference. 
Crucially, they can constitute an authoritative source of truth for certain applications where factuality is critical. 
This aspect is particularly relevant for large language models (LLMs)~\parencite{Yang2024-kgllm}, which are rapidly becoming part of the web experience for many users. 
Unfortunately, constructing large knowledge bases is costly, since population partly relies on manual data collection. 
This issue makes automated methods that can extract structured data from unstructured text particularly attractive when it comes to adding data to a knowledge base.

In natural language processing (NLP), knowledge graph extraction (KGE) refers to the task of extracting semantic triples of the form (subject, relation, object) from text to form a structured, directed graph of facts. 
A more advanced and realistic setting is temporal knowledge graph extraction (TKGE), where the temporal dimension of each fact is explicitly modeled by adding a timestamp, resulting in quadruples (subject, relation, object, timestamp). 
This formulation captures not only relationships but also their validity and evolution over time, enabling reasoning about sequences and durations of events, a crucial capability for representing dynamic real-world knowledge.

While static knowledge graphs treat all facts as timeless, most real-world knowledge is inherently temporal: events occur, evolve, and cease to hold. 
This temporal structure is especially critical in the era of LLMs.
Because LLMs are trained on massive but fixed textual corpora, they cannot easily be updated to reflect newly emerging or time-sensitive information. 
As a result, they often rely on external sources to stay current. 
In such scenarios, temporal knowledge graphs (TKGs) play a vital complementary role by providing explicit, temporally grounded knowledge that LLMs can query or condition on to answer on-time questions and handle realistic, temporally sensitive tasks such as event forecasting, dynamic question answering, or causal reasoning.
They serve as dynamic factual memory systems, providing continuously evolving context that static models alone cannot maintain.

By distinguishing between what was, is, and will be true, TKGs prevent LLMs from conflating outdated and current facts, mitigating hallucination risks in time-dependent reasoning. Unlike static graphs, which risk freezing knowledge in a timeless snapshot, TKGs continuously align with the evolving world, providing the temporal precision and factual consistency required for safe, interpretable, and trustworthy LLM-based systems.

The static KGE task is well-studied, and multiple datasets are available for training or evaluation, such as REBEL~\parencite{Cabot2021-rebel}, WebNLG~\parencite{Ferreira2020-webnlg} and Wiki-NRE~\parencite{Distiawan2019-wiki_nre}. 
Comparatively, resources are scarce for the temporal version of the task. 
While there is recent effort on that front~\parencite{Zhu2025-tkge_dataset}, the lack of resources limits the development of TKGE systems.

Concurrently, LLMs have been driving recent progress in natural language processing. 
LLM-based systems have achieved impressive results for many tasks, and TKGE is no exception to that success~\parencite{Zhang2024-edc}. 
Unfortunately, it is difficult to assess the real performance of these types of systems because of the possible overlap between their training and test datasets.
This issue, known as \textit{data contamination}, is both very common and difficult to detect and address due to the scale and opacity of current pre-training datasets and the black box aspect of models~\parencite{Dong2024-gen_vs_mem}. 
For TKGE, we hypothesize that this leads to inflated performance reporting on texts consisting of facts seen at training time.
Additionally, models can exhibit surprising biases when it comes to temporal reasoning in general.
For example, \textcite{Jain2023-timellm_bench} finds that LLMs struggle more when it comes to past rather than future events. 
In today's web landscape, LLMs are doubly important: they can be used to extract structured data as we already mentioned, but are also increasingly becoming an aspect of the mainstream web experience as they are easily accessible for the average user. Therefore, understanding their strengths and weaknesses for specific use cases is crucial.

Despite recent progress, assessing the true capabilities and biases of LLM-based TKGE systems remains challenging. Reliable, contamination-free benchmarks are essential for meaningful evaluation.
Existing datasets often overlap with the massive corpora used to pretrain LLMs, leading to artificially inflated results. 
Ideally, a TKGE benchmark should contain events unseen by the LLM under evaluation, occurring in a time period beyond the model's training horizon.

\begin{figure*}[tb]
  \centering
  \includegraphics[width=0.9\linewidth]{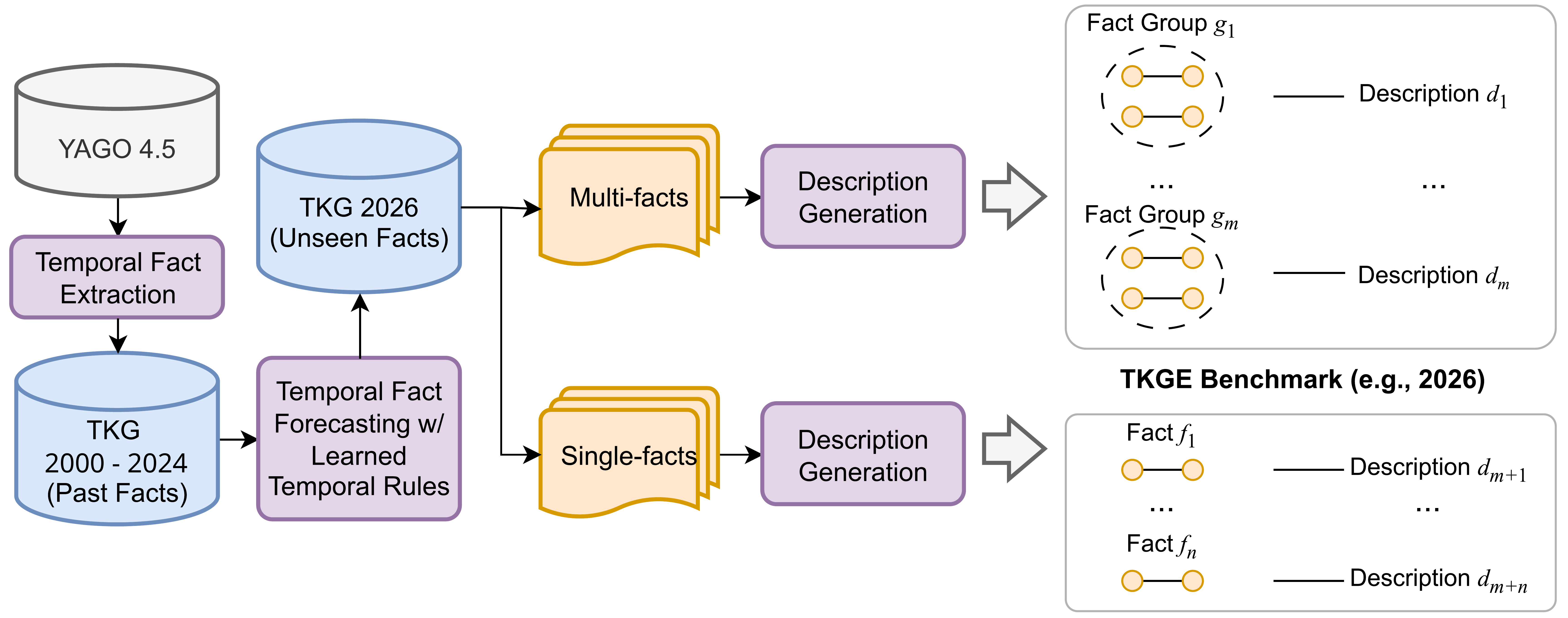}
  \caption{Overview of our methodology for generating Temporal Knowledge Graph Extraction (TKGE) benchmarks. Starting from historical temporal facts (1950–2024) extracted from YAGO 4.5, we forecast unseen future temporal facts (e.g., year 2026) using learned temporal rules (TLogic). The forecasted facts are grouped into single- and multi-fact scenarios, and LLMs generate corresponding textual descriptions. The resulting synthetic dataset serves as a renewable, contamination-resistant benchmark for evaluating TKGE systems.
  In essence, our method first predicts plausible future facts through rule-based temporal forecasting and then verbalizes them using an LLM, aligning symbolic and natural representations.}
  \Description{A diagram showing an overview of our methodology foe generating TKGE benchmarks.}
  \label{fig:dataset-generation}
\end{figure*}

To meet this need, we introduce a generalizable and renewable methodology for synthesizing unlimited future TKGE datasets composed of predicted, unseen temporal facts. Our approach is designed to serve as a long-term benchmark generation framework that remains valid even as new LLMs are developed.
Our methodology follows a two-stage generation process (illustrated in Figure~\ref{fig:dataset-generation}):
\begin{enumerate}
    \item \textbf{Temporal Forecasting:} Starting from an existing TKG, we employ a temporal knowledge graph forecasting (TKGF) system to generate plausible future facts. These predicted facts are then filtered according to the TKG schema to ensure structural validity.
    \item \textbf{Quadruple-to-Text Generation:} We leverage a LLM to verbalize each predicted fact into a natural textual description. The resulting dataset consists of future quadruples semantically aligned with their corresponding descriptions, suitable for use in TKGE evaluation.
\end{enumerate}

By generating facts that extend beyond the temporal horizon of all existing training corpora, our methodology inherently prevents data overlap with any past or present LLMs, yielding a contamination-free benchmark by design. Moreover, this framework is scalable and reproducible, allowing researchers to generate new datasets for future timeframes, ensuring sustainable and fair evaluation as LLMs evolve.
Our contributions are summarized as follows:
\begin{itemize}
    \item We propose a scalable and reproducible methodology for generating unlimited unseen future temporal datasets, establishing a foundation for fair, long-term, and reproducible evaluation.
    \item We provide a concrete instantiation of this methodology through the construction of the YAGO 2026 dataset, derived from the YAGO 4.5 knowledge base \parencite{Suchanek2024-yago4.5}.
    \item We produce an empirical confirmation of data contamination effects by benchmarking the state-of-the-art Extract, Define, and Canonicalize (EDC) framework \parencite{Zhang2024-edc}, demonstrating that LLM performance declines when evaluated on temporally unseen future facts. 
    \item We adapt the WebNLG 2020 KGE evaluation metrics \parencite{Ferreira2020-webnlg} to the temporal setting, and we release our code and datasets to foster reproducibility and future research.\footnote{\url{https://github.com/Aethor/fiction}}\footnote{\url{https://github.com/Aethor/edc}}
\end{itemize}

The rest of this paper is organized as follows. 
Section~\ref{sec:related-work} reviews related work, including TKGE datasets, temporal forecasting, and quadruple-to-text generation. 
Section~\ref{sec:method} introduces our dataset generation method, followed by Section~\ref{sec:yago-2026-dataset}, which presents the resulting dataset, YAGO 2026. 
Section~\ref{sec:measuring-td} reports experiments comparing the EDC framework's performance on YAGO 2026 versus a dataset of known facts, confirming the impact of data contamination. 
Finally, Section~\ref{sec:conclusion} concludes with insights on the long-term role of renewable temporal benchmarks in evaluating future LLM-based systems.

\section{Related Work}
\label{sec:related-work}
\subsection{Temporal Knowledge Base Extraction Datasets}
\label{sec:temp-knowl-base-cons}
KGE, the extraction of semantic triples from text, is a task that has garnered a lot of attention in NLP across the years~\parencite{Zhong2023-tkgc_survey,Ferreira2020-webnlg}. There are existing datasets to evaluate the task, such as WebNLG~\parencite{Ferreira2020-webnlg}, REBEL~\parencite{Cabot2021-rebel} and Wiki-NRE~\parencite{Distiawan2019-wiki_nre}. Comparatively, its temporal counterpart that necessitates the extraction of temporal quadruples is less studied. Even more problematic, aside from a the recent Tem-DocRED dataset proposed by \textcite{Zhu2025-tkge_dataset}, means of evaluation for this task are rare, and there are no standardized metrics available. While Tem-DocRED is useful, it only exhibits past facts, and may therefore suffer from the data contamination problem. 
As such, the reported performance of LLM-based systems evaluated on that dataset may be overly optimistic. By contrast, we propose a new synthetic dataset with future unseen facts. Our generation technique is reminiscent of existing KGE datasets such as REBEL or Wiki-NRE, although they were created by aligning text and triplets rather than using LLMs to generate descriptions.

\subsection{Temporal Knowledge Graph Forecasting}
\label{sec:temp-knowl-graph-fore}

To generate our synthetic dataset, we perform TKGF, which is the task to predict future temporal facts given an already known collection of facts. There are multiple classes of methods to solve that task. Several techniques are based on learning embeddings to compute a score function for a query. For example, tensor decomposition techniques like T-SimplE~\parencite{Lin2020-tsimple} or TNTComplex~\parencite{Lacroix2020-tntcomplex} represent a TKG as a 4D tensor and decompose it to obtain latent representations. Other works like HyTE~\parencite{Dasgupta2018-hyte} or ChronoR~\parencite{Sadeghian2021-chronor} regard timestamps as transformation of a static embedding. TimeTraveler~\parencite{Sun2021-timetraveler} uses reinforcement learning, framing the task as a problem of linking a subject entity to its object by navigating through an agent chosen path in the graph. Meanwhile, rule-learning methods such as TLogic~\parencite{Liu2022-tlogic}, TILP~\parencite{Xiong2023-tilp} or TEILP~\parencite{Xiong2024-teilp} learn rules by sampling random walks on the temporal graph, yielding more interpretable models. 
Some recent works~\parencite{Chang2025-tgl_llm,Lee2023-tkgf,Luo2024-coh,Liao2024-gentk} leverage LLMs for TKGF.

While recent LLM-based systems obtain state-of-the-art performance, we refrain from using them in our generation technique to prevent importing their bias into our evaluation dataset. The recent re-evaluation study by \textcite{Gastinger2024-tkgf_baseline} show that it is difficult to draw a conclusion concerning the best forecasting system in all situations. Therefore, we select the rule learning based method TLogic for its simplicity and its support for day-level event granularity. While the more recent TILP system is similar to TLogic, we do not select it as it does not support day-level granularity and focuses on events with time ranges (represented by a start and end timestamp), while we concern ourselves to singular events that can be represented with a simpler timestamp.

Finally, we note that forecasting performance is not strictly required for our application. The generated facts only needs to be plausible, but are not constrained to strict correctness. Additionally, by definition, there is no way to verify the correctness of a future fact at generation time.

\subsection{Quadruples-to-text and Dataset Generation}
\label{sec:quadruples-to-text}

Converting triples to text is a well-studied task in NLP. Notably, it has been part of the WebNLG challenge for a number of years~\parencite{Gardent2017-webnlg, Ferreira2020-webnlg,Cripwell2023-webnlg}. LLMs have been leveraged successfully, as the GPT-3.5 based submission of the DCU-NLG-PBN team at WebNLG 2023 shows~\parencite{Lorandi2023-webnlg_gpt35}. The triples-to-text task has also already been used to generate datasets: the K\textsc{E}LM corpus~\parencite{Agarwal2021-tekgen}, a collection of pairs of triples and text, is generated using a T5 model~\parencite{Raffel2019-t5} trained for triples-to-text generation. 
In our case, we perform quadruples-to-text using the Llama3.3-70B model~\parencite{Grattafiori2024-llama3} to generate descriptions.

\section{Unseen Temporal Knowledge Generation}
\label{sec:method}
Let $\mathcal{E}$ be the set of entities of a TKG $\mathcal{G}$, $\mathcal{R}$ its set of possible relationships and $\mathcal{T}$ its set of timestamps. $\mathcal{G}$ is composed of a set of quadruples of the form $q = (e_s, r, e_o, t)$, with $e_s,e_o \in \mathcal{E}$, $r \in \mathcal{R}$ and $t \in \mathcal{T}$. Each such quadruple denotes an edge of the graph, representing a relationship $r$ between subject entity $e_s$ and object entity $e_o$ at timestamp $t$. For a TKG $\mathcal{G}$, all timestamps in $\mathcal{T}$ are set within a defined time range $[t_s, t_e]$.

Given such a TKG, we wish to generate a TKGE benchmark with fictional, unseen temporal knowledge. 
To do so, we propose a two-step generation method. First, given the set of quadruples of the initial TKG $\mathcal{G}$, we generate a new TKG $\mathcal{G}'$ where all facts are set within a future time range $[t_{s'}, t_{e'}]$, with $t_{s'} > t_e$. We generate the new quadruples by sampling partial quadruples, and using the TLogic forecasting system~\parencite{Liu2022-tlogic} to complete them. Additionally, we make sure these quadruples are valid by filtering the ones that do not conform to the database schema. Then, we generate descriptions for these facts using an LLM. We propose a simple single-fact setup and a more realistic multi-facts setup. In the single-fact setup, we create examples as simple pairs, each composed of a quadruple and a description. We intend this setup to be used as a simple diagnostic to verify the extent of data contamination. Meanwhile, in the multi-facts case, each example has multiple quadruples and a single description from which the quadruples should be extracted. This setup allows to benchmark extraction system performance in a more realistic use-case. We present a summary of our generation method in Figure~\ref{fig:dataset-generation}.

\subsection{Source Data}
\label{sec:yago-datas-extr}

Before describing our dataset generation process, we start by discussing our choice of the initial TKG $\mathcal{G}$. While there are existing well known TKG datasets such as ICEWS14~\parencite{GarciaDuran2018-icews14}, ICEWS18~\parencite{Jin2020-icews18}, GDELT~\parencite{Leetaru2013-gdelt}, WIKI~\parencite{Leblay2018-wiki} or YAGO 3~\parencite{Mahdisoltani2013-yago3}, these are not the most up to date. Since we want to generate future facts, it is desirable to start the generation process from a more recent database to improve forecasting coherence. Therefore, we select YAGO 4.5~\parencite{Suchanek2024-yago4.5}, a recent cleaned-up subset of Wikidata which covers events up to 2023.

YAGO 4.5 weighs 124GB, but contains semantic triples without timestamps as well as temporal facts in the form of quadruples. We extract a TKG from YAGO 4.5, limiting ourselves to temporal facts from 2000 to 2024. We filter relations with a numeric or unique id as object (such as \texttt{populationNumber} or \texttt{isbn}). We also specifically filter the \texttt{affiliation} and \texttt{alumniOf} relations, since we found them to be easily confused or replaced with the \texttt{memberOf} relation. We filter rare entities by recursively removing entities with a degree of 1 as done by \textcite{Dasgupta2018-hyte}. In terms of timestamps, YAGO 4.5 has intervals of the form $[t_{start}, t_{end}]$, with $t_{start} \leq t_{end}$. 
Because our focus is on specific events rather than intervals, we perform \textit{linearization}: for a given fact $(e_s, r, e_o, [t_{start}, t_{end}])$, we create two facts $(e_s, r_{start}, e_o, t_{start})$ and $(e_s, r_{end}, e_o, t_{end})$. Here, $r_{start}$ represents a new relation that denotes the start of relation $r$, while $r_{end}$ is similar for the end of relation $r$. As an example, we linearize the fact \texttt{(Bill Gates, memberOf, Boston University Terriers men's Basketball, [01-01-1958, 01-01-1959])} as the two facts:

\begin{itemize}
\item \texttt{(Bill Gates, startMemberOf, Boston University Terriers men's Basketball, 01-01-1958)}
\item \texttt{(Bill Gates, endMemberOf, Boston University Terriers men's Basketball, 01-01-1959)}
\end{itemize}

This transformation allows forecasting systems to handle complex temporal intervals through point-based reasoning, while preserving temporal semantics.
We exclude certain relationships from linearization when semantically inappropriate.  
For example, the \texttt{children} relation logically cannot have a $t_{end}$ timestamp, as it is not possible to stop being the child of someone.

\begin{figure*}[tb]
  \centering
  \includegraphics[width=\linewidth]{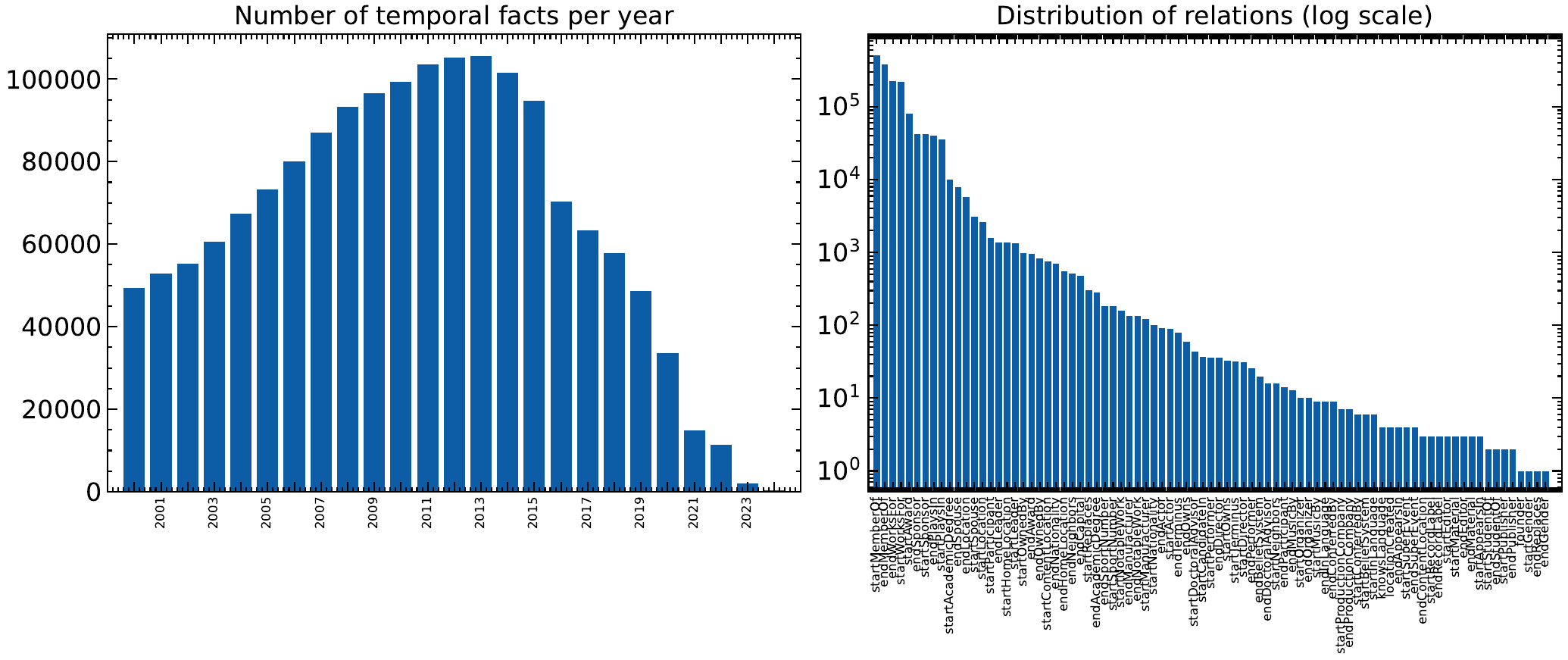}
  \caption{Statistics of the extracted YAGO 4.5 Temporal Knowledge Graph (TKG).
The left histogram shows the number of temporal facts per year, ranging from 1950 to 2023, showing a steady growth in recorded events until 2015, followed by a sharp decline in more recent years. The right histogram (log scale) presents the distribution of relations across all temporal facts, revealing a long-tailed pattern dominated by a few highly frequent relations such as startMemberOf, while most relations appear far less frequently. These distributions reflect both the temporal coverage and relational diversity of the extracted YAGO 4.5 TKG.}
  \Description{Two histograms. The first one shows the number of temporal facts per year in our extracted YAGO 4.5 TKG. The second one shows the number of facts per relation, the most frequent relation being \texttt{startMemberOf}.}
  \label{fig:yago4.5}
\end{figure*}

The resulting TKG has $1,627,497$ facts, $468,885$ distinct entities and $83$ different relations. We show its number of facts per year and relation distribution in Figure~\ref{fig:yago4.5}. 
The peak of number of facts occur in 2013 with approximately $100,000$ facts, and steadily decreases afterwards. 
We observe that the relation distribution is far from uniform, containing extremely frequent and rare relations.

\subsection{Future Facts Generation using TLogic}
\label{sec:new-facts-generation}

\subsubsection{Background on TLogic}
\label{sec:background-tlogic}

TLogic~\parencite{Liu2022-tlogic} is a TKGF system based on temporal rule mining. A temporal rule has the following form:

\begin{equation}
  \label{eq:trule-def}
  (E_1, r_h, E_{l+1}, T_{l+1}) \leftarrow \bigwedge_{i=1}^{l}{(E_i, r_i, E_{i+1}, T_{i+1})},
\end{equation}

With $r_i \in \mathcal{R}$, $E_i$, $T_i$ standing as entity and timestamp variables, and the temporal constraint $T_{l+1} \geq T_{l} \geq ... \geq T_1$. The leftmost part is said to be the \textit{rule head}, while the rightmost part is said to be the \textit{rule body}. The \textit{rule head} is said to be true when its body holds true. For example, a possible rule is:

\begin{equation}
  \label{eq:trule-example}
  \begin{aligned}
    (E_1, endLeader, E_3, T_3) & \leftarrow (E_1, endLeader, E_2, T_1)\\
                               & \wedge (E_2, endWorksFor, E_1, T_2)\\
                               & \wedge (E_1, startLeader, E_3, T_3)
  \end{aligned}
\end{equation}

To learn rules, TLogic samples an edge $(e_1, r_h, e_{l+1}, t_{l+1})$ corresponding to a rule head. Then, the system samples non-increasing temporal random walks of length $l+1$, representing a path backwards in time from entity $e_{l+1}$ to entity $e_1$. These walks are of the form:

\begin{equation}
  \label{eq:twalk}
  (e_{l+1}, r_l, e_l, t_l), (e_l, r_{l-1}, e_{l-1}, t_{l-1}), ..., (e_2, r_1, e_1, t_1),
\end{equation}

with $t_l \geq t_{l-1} \geq ... \geq t_1$. At each step, the transition distribution depends on timestamps, and is either uniform or exponentially weighted. A temporal walk is transformed into a cyclic temporal rule by replacing entities and timestamps with variables: the starting edge $(e_1, r_h, e_{l+1}, t_{l+1})$ becomes the rule head, and the temporal walk becomes its body. Each rule has an associated confidence between 0 and 1, which is the ratio of the rule support and the body support. The body support is defined as the number of occurrences of the body in the TKG, while the rule support is the number of time the rule holds true when the body occurs.

To apply rules and answer a query of the form $(e_s, r, ?, t)$, TLogic iterates over rules learned for relation $r$. For each of these rules, it samples all body groundings, and retrieve candidate objects and confidence values from these groundings.

\subsubsection{Future Facts Generation Algorithm}
\label{sec:future-facts-gen_alg}

\begin{algorithm}[tb]
  \caption{Future facts generation algorithm}
  \label{alg:fact-gen}
  \textbf{Input}:
  \begin{itemize}
    \item day-level timestamp $t$
    \item set of past quadruples $\mathcal{Q}$
    \item set of subjects $\mathcal{S}$
    \item set of relations $\mathcal{R}$
    \item relations sample weights $\mathcal{W}_y$ for past year $y$
    \item number of subjects to sample $k$
    \item database schema $schema$
  \end{itemize}
  \textbf{Output}:
  \begin{itemize}
    \item future fact $q'$
  \end{itemize}
  \begin{algorithmic}[1] 
    \STATE $q' \leftarrow \varnothing$
    \WHILE{$q' = \varnothing$}
        \STATE $r \leftarrow sample\_weighted(\mathcal{R}, 1, \mathcal{W}_y)$
        \STATE $\mathcal{S}_r \leftarrow \{e_s \in S : allowed(e_s, r, schema)\}$
        \STATE $\mathcal{S}_r \leftarrow sample\_uniform(\mathcal{S}_r, k)$
        \STATE $\mathcal{Q}' \leftarrow \emptyset$
        \FOR{$e_s$ in $S_r$}
            \STATE $\mathcal{Q}' \leftarrow \mathcal{Q}' \cup TLogic(\mathcal{Q}, (e_s, r, ?, t))$
        \ENDFOR
        \STATE $\mathcal{Q}' \leftarrow \{q \in \mathcal{Q}' : valid(q, schema)\}$
        \STATE $q' \leftarrow max_{confidence}(\mathcal{Q}')$
    \ENDWHILE
    \RETURN $q'$
  \end{algorithmic}
\end{algorithm}

Our future facts generation algorithm, presented in details in Algorithm~\ref{alg:fact-gen}, works as follows. For a given timestamp $t$ at the day level, we wish to generate a future quadruple $q'$. TLogic can answer a query of the form $(e_s, r, ?, t)$, yielding us a confidence-ranked set of object entities that we can use to complete the input quadruple. Thus, we need to sample a subject $e_s$ and a relation $r$ before querying TLogic. Since we wish to mimic the frequency of relations in YAGO 4.5 to obtain a somewhat realistic dataset, we start by sampling a relationship $r$, using the frequency of relations observed from a recent year of YAGO 4.5 as weights.\footnote{In practice, we use 2022 as our relation frequency guide. As YAGO 4.5 was released during 2023, we guess this is the most recent year with sufficiently complete data.} Given that relation $r$, we sample a set $S_r$ of $k$ subjects among the allowed subjects according to the database schema. $k$ is a parameter that improves the quality of the dataset by increasing the number of generated quadruples to choose from, but increases generation time. In practice, we set $k = 4$. Finally, we query TLogic with the set of queries $\{(e_s, r, ?, t), e_s \in S_r\}$, and return the valid generated quadruple with the highest confidence.

Filtering invalid facts is an important part of the algorithm. While TLogic is able to infer new facts, these facts do not necessarily conform to the database schema: for example, the quadruple \textit{(Taipei, startDoctoralAdvisor, Linus Torvalds, 2026-01-01)} could be generated by TLogic, but is problematic as it stands to reason that a city cannot be a doctoral advisor. Therefore, we leverage YAGO's schema to filter the generated facts, keeping only valid facts according the schema. We define three filtering mechanisms:

\begin{itemize}
\item \textbf{Subject pre-filtering}: Before querying TLogic, given a relation $r$, we pre-filter subjects by eliminating those that cannot have $r$ as a property. 
\item \textbf{Coherency pre-filtering}: Since we linearize facts and many relations are exclusive (e.g. most of the time, an individual works at a single place at the same time), we also maintain coherency by disallowing opening an already opened relation and closing a non open relation. Given a subject $e_s$, we define an open relationship for subject $e_s$ as the existence of a starting quadruple $(e_s, r_{start}, e_o, t_{start})$ and the non-existence of the corresponding ending quadruple $(e_s, r_{end}, e_o, t_{end})$, where $t_{end} \geq t_{start}$. Conversely, we simply define a closed relationship as a non-open one.
\item \textbf{Object filtering}: After the query, we filter quadruples if the type of their object is incompatible with relation $r$. 
\end{itemize}

While we could theoretically use our pre-filtering steps \textit{after} querying TLogic, we apply them first as an optimization step to avoid running unnecessary queries.

Since we wish to generate facts over a time range $[t_{s'}, t_{e'}]$ specified at a day-level granularity, we start by generating facts for starting day $t_{s'}$, and iteratively apply our algorithm for each subsequent day by updating the set of past quadruples $\mathcal{Q}$ with our newly generated facts at each step.

\subsection{Fact Descriptions Generation}
\label{sec:fact-descr-gener}

Given a set of facts, we generate descriptions for them using Llama3.3-70b~\parencite{Grattafiori2024-llama3}.\footnote{The model  accessed through the Vertex AI API: \texttt{llama-3.3-70b-instruct-maas}}
These descriptions can then be used as input when evaluating the TKGE task, while quadruples can be used as reference. We propose two different setups: a simple single-fact diagnostic setup, where a system has to extract a single quadruple from a short text, and a more realistic multi-facts setup where multiple facts have to be extracted from the same text. We give the details of our generation prompts in Appendix~\ref{sec:fact-desc-prompt}.

\subsubsection{Single-Fact Setup}
\label{sec:single-fact-setup}

In the single-fact setup, we generate short texts containing a single temporal fact. Given a fact $(e_s, r, e_o, t)$, we prompt the LLM to generate a short text containing that fact, in a newspaper style. To improve diversity in our dataset, we randomly sample different styles for the timestamp $t$ in the prompt. We also prompt the LLM to generate a headline with a fictional date of publication for our document in 25\% of cases, to increase task difficulty. 

\subsubsection{Multi-Facts Setup}
\label{sec:multi-facts-setup}

In the multi-facts setup, we generate slightly longer texts containing 2 to 4 temporal facts. We hypothesize that a news article is likely more plausible if it has a general theme or talks about related entities: Therefore, we propose to group similar facts together for generation. We assess the similarity of facts using the proximity between their entities and timestamps, and use agglomerative clustering to create clusters. Let $q = (e_s, r, e_o, t)$ and $q' = (e_s', r', e_o', t')$ be two quadruples. We compute their distance using function $d$:

\begin{equation}
  \label{eq:fact-dist}
  d(q, q') = \alpha \frac{(d_{ent}(e_s, e_s') + d_{ent}(e_o, e_o'))}{2} + (1 - \alpha) d_{ts}(t, t'),
\end{equation}

\noindent where $d_{ent}$ is a function measuring the distance between entities, and $d_{ts}$ measures the temporal distance between two timestamps. To compute $d_{ent}$, we first construct the inheritance graph $H$ of YAGO entity classes. Let $C$ and $C'$ be the set of classes of entity $e_*$ and $e_*'$ respectively, and $d_H(c, c')$ be the number of links in the shortest path between classes $c$ and $c'$ in $H$. We compute $d_{ent}$ as:

\begin{equation}
  \label{eq:ent-dist}
  d_{ent}(e_*, e_*') = 1 - \frac{1}{\displaystyle \min_{c \in C, c' \in C'}(d_H(c, c')) + 1}.
\end{equation}

\noindent We compute the temporal distance $d_{ts}$ between facts using a logistic function with a midpoint of half a year:

\begin{equation}
  \label{eq:ts-dist}
  d_{ts}(t, t') = \frac{1}{1 + \mathrm{e}^{-\kappa (|t' - t| - \frac{365}{2})}}.
\end{equation}

\begin{figure}
    \centering
    \includegraphics[width=1.0\linewidth]{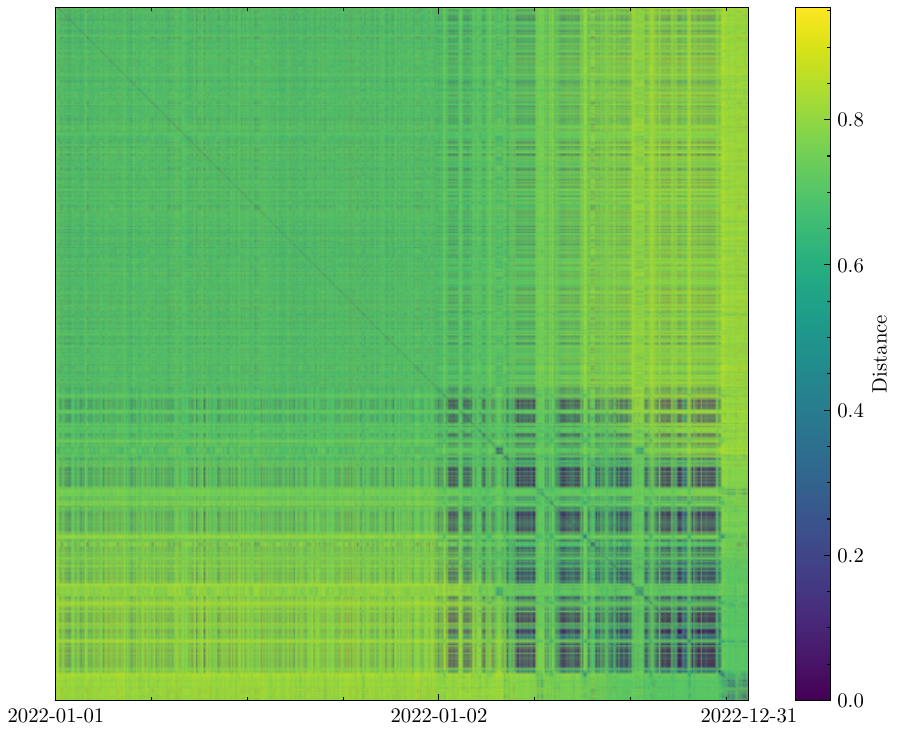}
    \caption{Distance matrix for the facts present in the 2022 year of YAGO. Most facts occur in the first day of the year, likely because the date for some events is not precise and set to this day by default. Blocks generally indicate series of concurrent related facts, such as a decoration ceremony where many people receive similar awards.}
    \label{fig:dist_matrix}
\end{figure}

\noindent In all experiments, we empirically set $\alpha = 0.8$ and $\kappa = 0.03$. Figure~\ref{fig:dist_matrix} shows the distance matrix for the facts of the 2022 year of YAGO.

\section{Generated Dataset}
\label{sec:yago-2026-dataset}

\subsection{Generated Facts}
\label{sec:generated-facts}

\begin{figure}[tb]
  \centering
  \includegraphics[width=\linewidth]{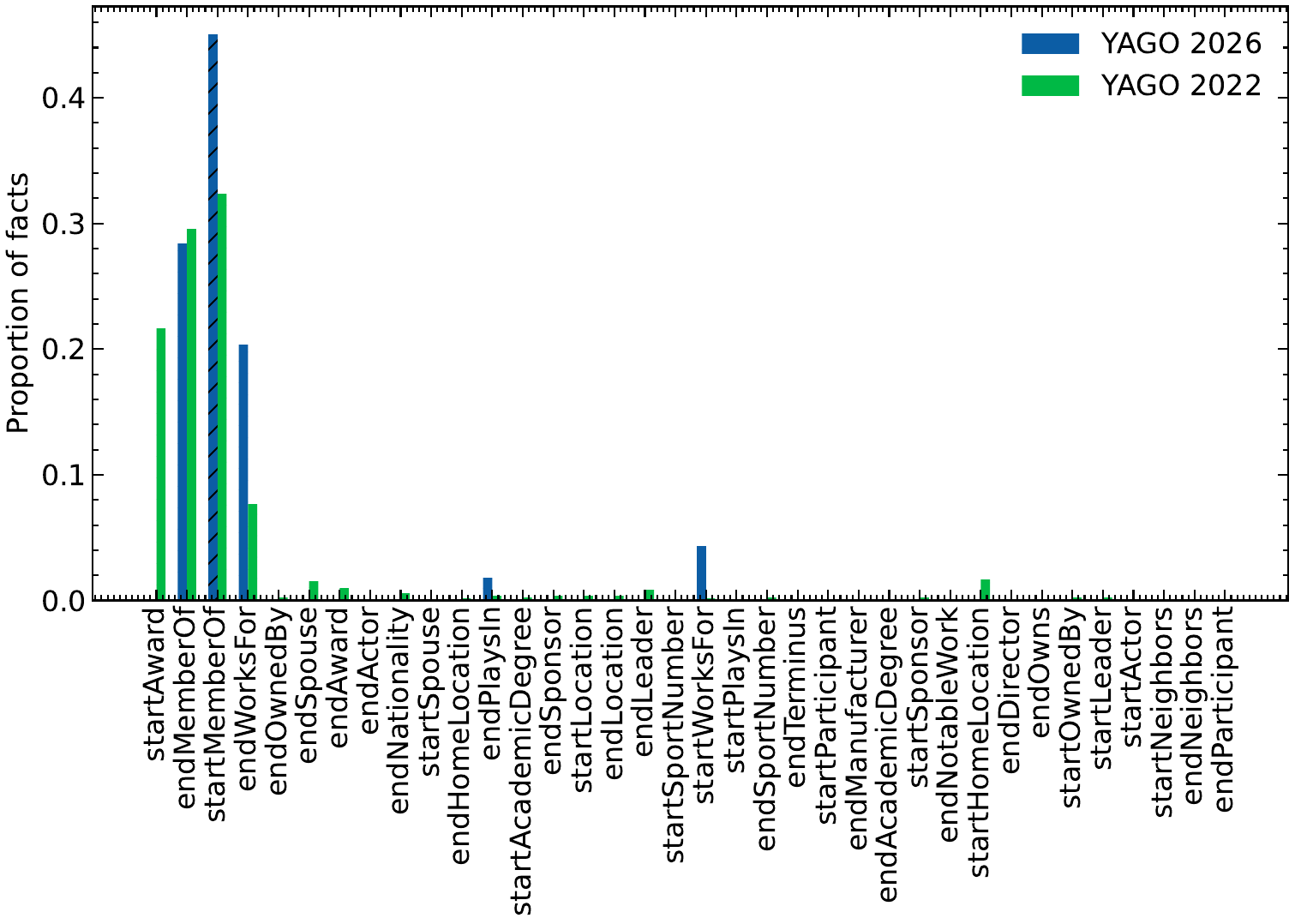}
  \caption{Proportion of facts per relation in the YAGO 2022 and YAGO 2026 datasets. 
  The fictional YAGO 2026 temporal knowledge graph is generated based on YAGO 2022. 
  For YAGO 2022, the top three relations are endMemberOf, endWorksFor, and startMemberOf. For YAGO 2026, the most frequent relations are startMemberOf, endMemberOf, and startAward. The distributions differ notably between the two datasets, reflecting the changes introduced by the temporal forecasting process.}
  \Description{A histogram, showing the proportion of facts for each relation in the YAGO 2022 and YAGO 2026 datasets. For YAGO 2022, the top 3 relations are \texttt{endMemberOf}, \texttt{endWorksFor} and \texttt{startMemberOf}. For YAGO 2026, these are \texttt{startMemberOf}, \texttt{endMemberOf} and \texttt{startAward}. The distribution is different between the two datasets.}
  \label{fig:2022_2026_rel_dist}
\end{figure}

As an example application of our methodology, we generate the \textit{YAGO 2026 TKGE} dataset for the period $[\text{2026-01-01}, \text{2026-12-31}]$, highlighting the practical capability of our approach to continuously produce future evaluation benchmarks.
We train the TLogic TKGF system using the 98\% earliest data of the YAGO 4.5 TKG, keep the latest 2\% as a test set for estimating forecasting performance, and use the full dataset as a base for forecasting. We extract rules of length of 1, 2 and 3, set the number of random walks to 200, and use the exponential transition distribution.

We report the performance of TLogic on the test set using time-aware filtering following other works~\parencite{Gastinger2022-tkgf_eval, Liu2022-tlogic}. At inference time, we use a time window of $2,048$ for performance reasons. We measure the performance of TLogic using Hits@10 (the proportion of queries for which at least one correct item is in the top 10 results) and Mean Reciprocal Rank (MRR, the mean of the reciprocals of the rank of the first correct item in the list of results). After training, the forecasting performance of TLogic on the test set is $43.09$ Hits@10 and $35.28$ MRR, which indicate moderate performance.

For each day of 2026, we generate the same number of facts as the corresponding day of 2022 in the YAGO TKG (with an upper limit of $m = 128$ facts to limit costs). In total, we generate $4,209$ new facts, with $18$ different relations. Figure~\ref{fig:2022_2026_rel_dist} shows the proportion of facts per relation in YAGO 2022 and YAGO 2026. We observe that, while we conceived Algorithm~\ref{alg:fact-gen} to mimic the relation distribution of YAGO 2022, the output distribution is not equivalent. 
This occurs because TLogic may not generate answers if applicable rules are absent.
Since there are not many rules for some relations (in particular rare ones), this leads to the algorithm not being able to exactly fulfill the target distribution. We experimented with variants of this algorithm, and this version is the one that gave us the best results regarding the distribution.

\subsection{Generated Descriptions}
\label{sec:gener-descr}

\begin{table*}[tb]
  \centering
  \begin{tabular}{p{8cm} p{9cm}}
    \toprule
    \textbf{Quadruple(s)} & \textbf{Generated Description} \\
    \midrule
    (Jacques Jones, startMemberOf, K.V.C. Westerlo, 2026-01-11) & \textit{``Sun, January 11th, 2026: Breaking News - Jacques Jones Joins K. V. C. Westerlo. Jacques Jones started being a member of K. V. C. Westerlo today, January 11th, 2026.''} \\
    \midrule
    (Patrik Klačan, endMemberOf, Slovakia National under 15 Football Team, 2026-01-06)
    
    (Pavel Vieira, startMemberOf, Çaykur Rizespor, 2026-01-07)
    
    (Mark Kitching, endMemberOf, Rochdale A. F. C., 2026-01-07)
    
    (Matthew Hockley, endMemberOf, Truro City F. C., 2026-01-10) & \textit{``Friday, 2026-01-02: Mid-Week Sports Updates. Patrik Klačan's tenure with the Slovakia national under-15 football team came to an end on Tuesday, January 6th, 2026. The following day, Wednesday, January 7th, 2026, Pavel Vieira joined Çaykur Rizespor as a new member. Also on January 7th, 2026, Mark Kitching parted ways with Rochdale A. F. C. Later, on Saturday, January 10th, 2026, Matthew Hockley ended his membership with Truro City F. C.''} \\
    \bottomrule
  \end{tabular}
  \caption{ Samples of automatically generated descriptions.
Each row shows one or more temporal knowledge quadruples alongside the corresponding text generated by our LLM-based description generation step. The first example illustrates the single-fact setup, where the model produces a concise, coherent description for a single temporal fact. The second example demonstrates the multi-facts setup, in which multiple related facts are combined into a single narrative resembling a short news report. The generated texts accurately reflect the entities, relations, and timestamps of the underlying quadruples, while maintaining natural linguistic fluency and temporal coherence. These samples illustrate the ability of our approach to transform structured temporal knowledge into realistic and contextually grounded textual descriptions suitable for temporal knowledge graph extraction benchmarking.}
  \label{tab:generated-facts}
\end{table*}

Table~\ref{tab:generated-facts} shows samples of generated descriptions. In total, we generate $4,209$ single-fact descriptions and $852$ multi-fact descriptions, for a total cost of $9.88$\$.  We initially experimented with Llama3.1-8b instead of Llama3.3-70b, but found descriptions were of noticeably lower quality. In particular, the model would often fail at correctly representing the order of events, incorrectly using past or future tense, and would sometimes fail at including the quadruple timestamp in its description. To ensure quality on the generated descriptions, we perform a small manual study on 30 single-fact descriptions and 50 multi-facts descriptions. We survey three different quality aspects: coverage, relevance and self-consistency. Coverage verifies that all facts are included in the generated description. Relevance verifies whether all included facts are relevant. Finally, self-consistency checks if the description is logically consistent. For single facts, due to the simplicity of the setup, we do not find any issues. For the multi-facts setup, we find perfect coverage in 49/50 cases, perfect relevance in 48/50 cases and perfect self-consistency in 41/50 cases, which shows the high accuracy of generated descriptions.

\section{Measuring Data Contamination}
\label{sec:measuring-td}

In this section, we perform experiments to highlight the interest of our dataset. In particular, we demonstrate how it can quantify the extent of data contamination in an LLM-based TKGE system.

\subsection{Method}
\label{sec:method-td}

We benchmark the recent state-of-the-art \textit{Extract, Define, Canonicalize} (EDC) framework~\parencite{Zhang2024-edc}, based on LLMs. We compare the performance of EDC on our YAGO 2026 dataset, and on a similar dataset extracted from the year 2022 of YAGO. 
Our hypothesis is that, due to memorization effects, the extraction system is expected to perform worse on the YAGO 2026 dataset since it is composed of unseen facts. 
Since the original EDC implementation only considers triple extraction, we adapt it for quadruple extraction by modifying its implementation and adjusting prompts. The three phases of EDC are as follows:

\begin{enumerate}
\item \textbf{Extract}: Prompt a LLM to extract quadruples in an open fashion. At this stage, the extracted relations might not be valid.
\item \textbf{Define}: For each extracted relation, prompt a LLM to generate a description.
\item \textbf{Canonicalize}: Finally, map each extracted relation to the closest canonical relation. This is done by computing the proximity between descriptions generated in the \textit{Define} step and descriptions for canonical dataset relations, using an embedding model such as SentenceBERT~\parencite{Reimers2019-sbert}.
\end{enumerate}

In our experiments, we reproduce one of the setup used in the original EDC article. We use Mistral-7B-Instruct-v0.2~\parencite{Jiang2023-mistral7b} for the \textit{Extract} phase, GPT3.5-turbo~\parencite{Brown2020-chatgpt} for the \textit{Define} and \textit{Canonicalize} phases,\footnote{While it is not explicit in the original EDC article, GPT3.5-turbo is used in the \textit{Define} and \textit{Canonicalize} phases for all setups.} and the E5-mistral-7b-instruct model~\parencite{Wang2024-e5} as embedder for the \textit{Canonicalize} phase. We also report results using the Mistral-7B-Instruct-v0.2 model alone as a baseline.

\subsubsection{Datasets}
\label{sec:method-data}

We compare the performance of our version of EDC on our YAGO 2026 dataset, and on a dataset we extract from YAGO for the year 2022 (YAGO 2022).  To extract this 2022 dataset, we 1) keep YAGO 4.5 temporal facts for 2022; 2) generate description for these facts using Llama3.3-70B as described in Section~\ref{sec:fact-descr-gener}.

To ensure a fair comparison, we resample both datasets in order for them to have 1) a similar relation distribution and 2) a similar timestamp distribution across the surveyed year. This setup guarantees that the difference between YAGO 2022 and YAGO 2026 is strictly caused by their underlying facts.


\subsection{Evaluation}
\label{sec:evaluation-td}

To our knowledge, there are no standard metrics when it comes to quadruple extraction. Therefore, we adapt the metric used for the WebNLG 2020 triple extraction challenge to quadruple extraction~\parencite{Ferreira2020-webnlg}. The original metric has 4 different modes: \textit{strict}, \textit{exact}, \textit{type} and \textit{partial}. In \textit{strict} mode, all elements of the candidate and reference quadruple should be the same. In \textit{exact} mode, the type two elements is not required to match, but the element strings themselves should be exactly identical. In \textit{partial} mode, the strings do not have to be identical. Finally, in \textit{type} mode, if a candidate element has a partial match with a reference element and they are of the same type, it is considered as a full match. We extend this logic to quadruples. We also correct a few shortcomings of the metric, as explained in more details in Appendix~\ref{sec:quadr-extr-eval}.

\subsection{Results}
\label{sec:results-td}

\begin{table*}[tb]
  \centering
  \begin{tabular}{lllllllll}
    \toprule
    \textbf{Dataset}   & \textbf{Strict} & \textbf{Exact} & \textbf{Type} & \textbf{Partial} & \textbf{Strict} & \textbf{Exact} & \textbf{Type} & \textbf{Partial} \\
    \textbf{System}    &            \multicolumn{4}{c}{EDC+Mistral-7B-Instruct-v0.2} & \multicolumn{4}{c}{Mistral-7B-Instruct-v0.2} \\
    \midrule
    YAGO 2022 (single) & 72.90**         & 73.97**        & 78.98**       & 76.39**          & 67.87** & 70.88** & 74.93** & 73.75** \\
    YAGO 2026 (single) & 70.08           & 71.12          & 77.08         & 73.67            & 61.88 & 64.60 & 69.60 & 67.53 \\
    \midrule
    YAGO 2022 (multi)  & 85.29           & 87.25*         & 93.63         & 90.68**          & 84.15 & 86.22* & 92.93 & 90.07** \\
    YAGO 2026 (multi)  & 84.30           & 85.15          & 93.39         & 88.67            & 83.01 & 84.22  & 92.83 &  88.16  \\
    \bottomrule
  \end{tabular}
  \caption{F1-score of TKGE systems on YAGO 2022 and YAGO 2026 for the single-fact and multi-facts setups. A permutation test shows that, in the single fact setup, results on YAGO 2022 are better in a statistically significant way (**: $p < 0.01$, *: $p < 0.05$)}
  \label{tab:single-results}
\end{table*}

We present results on our YAGO 2022 and YAGO 2026 datasets in Table~\ref{tab:single-results}. The EDC TKGE framework outperforms the baseline using solely Mistral-7b-Instruct-v0.2. Interestingly, scores on the multi-facts setup is considerably higher than on the single-fact setup, which might be explained because the number of context clues in this setup is higher, easing the extraction for the model. Overall, performance on the dataset of unknown future facts is lower than on the 2022 dataset of known facts in the single fact setup, with a statistically significant difference of 2.82 strict F1 using EDC. While we also observe this phenomenon on the multi-facts setup, the difference is less important (0.99 strict F1 using EDC) and not statistically significant depending on the metric, perhaps due to the easier nature of the setup. These results tend to show that the Mistral-7B-Instruct-v0.2 model suffers from a bias towards seen facts.

\subsubsection{Influence of Timestamps}
\label{sec:influence-timestamps}

\begin{table*}[tb]
  \centering
  \begin{tabular}{lllllllll}
    \toprule
    \textbf{Dataset}               & \textbf{Strict} & \textbf{Exact} & \textbf{Type} & \textbf{Partial} & \textbf{Strict} & \textbf{Exact} & \textbf{Type} & \textbf{Partial} \\
    \textbf{System}                &            \multicolumn{4}{c}{EDC+Mistral-7B-Instruct} & \multicolumn{4}{c}{Mistral-7B-Instruct} \\
    \midrule
    YAGO 2022 (single)                      & 72.90           & 73.97          & 78.98         & 76.39            & 67.87** & 70.88** & 74.93** & 73.75** \\
    YAGO 2022 (single) $\Rightarrow$ 2026   & 72.16           & 73.09          & 78.21         & 75.50            & 65.85   & 67.75   & 72.02   & 70.23   \\
    \midrule
    YAGO 2026 (single)             & 70.08           & 71.12          & 77.08         & 73.67            & 61.88   & 64.60   & 69.60   & 67.53   \\
    YAGO 2026 (single) $\Rightarrow$ 2022   & 70.45           & 71.71          & 77.50         & 74.29            & 64.72** & 66.70** & 72.69** & 69.71** \\
    \bottomrule
  \end{tabular}
  \caption{F1-score of EDC on YAGO 2022 and YAGO 2026 in the single-fact setup, when modifying the year of the timestamps. In the YAGO 2022 $\Rightarrow$ 2026 setting, we change the years in YAGO 2022 timestamps to 2026. We do the reverse for YAGO 2026 $\Rightarrow$ 2022. Differences are statistically significant for the Mistral-7B-Instruct-v0.2 baseline under a permutation test (**: $p < 0.01$).}
  \label{tab:ts-results}
\end{table*}

Our main hypothesis is that unseen events are more difficult to extract for LLMs. However, there is another factor that may increase the difficulty of the task for LLMs: timestamps. 
For example, \textcite{Tan2023-tempreason} find that LLM reasoning abilities are biased towards contemporary years. We are therefore interested in knowing how much timestamps alone influence our results. To measure that effect, we propose to start from our YAGO 2022 dataset, to change all timestamps to 2026 and to repeat our previous extraction experiments (YAGO 2022 $\Rightarrow$ YAGO 2026). Conversely, we also modify our YAGO 2026 dataset with 2022 timestamps to measure whether this renders the task easier (YAGO 2026 $\Rightarrow$ YAGO 2022).

As seen in Table~\ref{tab:ts-results}, we find that extraction for future timestamps is rendered more difficult. While this difference is not statistically significant using the EDC framework, the baseline model alone is more affected, with a significant difference of $-2.02$ strict F1. Meanwhile, extracting fictional events with past timestamps is made easier, with a significant improvement of $+2.84$ \textit{strict} F1 for the baseline model. These effects are absent from the multi-facts setup, as we observe in Appendix~\ref{sec:influence-timestamps-multi}.

Overall, our results tend to show that Mistral-7b-Instruct-v0.2 suffers from a timestamp bias, but that the EDC extraction framework is able to mitigate this issue. However, this timestamp bias alone cannot explain the larger difference we observe when comparing performance on the YAGO 2022 and YAGO 2026 datasets. The rest of that difference may therefore be attributed to the unseen nature of facts in YAGO 2026.

\section{Conclusion}
\label{sec:conclusion}
We presented a novel, reproducible methodology for generating unlimited synthetic TKGE datasets comprising unseen future temporal facts and their textual descriptions. As a practical demonstration, we introduced the YAGO 2026 dataset, designed as a long-term, contamination-resistant benchmark for evaluating LLM-based TKGE systems. Our experiments on the Mistral-7b-Instruct-v0.2 model tend to confirm data contamination biases, and highlight that LLMs can exhibit temporal biases toward timestamps prevalent in their training data~\parencite{Tan2023-tempreason}.

Limitations of our work include the inability of TLogic to generate facts for rare relationships due to sparse training data, and its indirect support for time ranges. Future work could explore advanced forecasting methods like TILP \parencite{Xiong2023-tilp}, extending them to finer day-level granularity. Furthermore, our analysis may be extended to a broader range of LLMs.

We note that the absence of standardized temporal-aware metrics remains a challenge for fair TKGE evaluation. We advocate for the development of standardized temporal-aware metrics to better assess temporal extraction accuracy. 

In the long term, we envision our approach forming the foundation for hybrid systems in which LLMs dynamically query evolving TKGs, bridging static pretrained models with real-time, temporally grounded reasoning.

\section{Ethics Statement}
This research relies exclusively on the publicly available YAGO knowledge base, which aggregates structured factual data from Wikipedia and Wikidata. 
These sources are openly licensed, ensuring that the data used in this work raise no copyright concerns.
No private, proprietary, or restricted materials were employed. 
LLMs were used exclusively for synthetic data generation and language refinement during manuscript preparation.  
The authors bear full responsibility for the content of this work.
\printbibliography

\appendix

\section{Influence of Timestamps: Multi-Facts Setup}
\label{sec:influence-timestamps-multi}
In this section, we present the results of our timestamp changing experiment in the multi-facts setup. Results can be found in Table~\ref{tab:ts-results-multi}. We observe small differences between setups: using EDC, performance slightly increases when changing timestamps, both in the YAGO 2022 $\Rightarrow$ 2026 and YAGO 2026 $\Rightarrow$ 2022 setups. Using the baseline Mistral-7B-Instruct-v0.2 model alone, we observe a small negative influence of updating timestamps to 2026, and a small positive influence of updating timestamps to 2022. However, we find under a randomized permutation test that these small differences are not statistically significant. Our hypothesis is that since the task in the multi-facts setup is easier compared to the single-fact setup, as we explain in the main text of the article, the influence of timestamps on the final score is greatly reduced.

\begin{table*}[t]
  \centering
  \begin{tabular}{lllllllll}
    \toprule
    \textbf{Dataset}   & \textbf{Strict} & \textbf{Exact} & \textbf{Type} & \textbf{Partial} & \textbf{Strict} & \textbf{Exact} & \textbf{Type} & \textbf{Partial} \\
    \textbf{System}    &            \multicolumn{4}{c}{EDC+Mistral-7B-Instruct-v0.2} & \multicolumn{4}{c}{Mistral-7B-Instruct-v0.2} \\
    \midrule
    YAGO 2022 (multi)                    & 85.59 & 87.25 & 93.63 & 90.68 & 84.15           & 86.22          & 92.93         & 90.07  \\
    YAGO 2022 (multi) $\Rightarrow$ 2026 & 86.06 & 87.44 & 94.23 & 90.93 & 83.12           & 84.95          & 92.47        & 89.07   \\
    \midrule
    YAGO 2026 (multi)                    & 84.30 & 85.15 & 93.39 & 88.67 & 83.01           & 84.22          & 92.93         & 88.16  \\
    YAGO 2026 (multi) $\Rightarrow$ 2022 & 84.47 & 85.54 & 93.42 & 89.10 & 83.06           & 84.29          & 92.88         & 88.28  \\
    \bottomrule
  \end{tabular}
  \caption{F1-score of EDC on YAGO 2022 and YAGO 2026 in the multi-facts setup, when modifying the year of the timestamps. In the YAGO 2022 $\Rightarrow$ 2026 setting, we change the years in YAGO 2022 timestamps to 2026. We do the reverse for YAGO 2026 $\Rightarrow$ 2022. We observe small differences that are not statistically significant under a permutation test.}
  \label{tab:ts-results-multi}
\end{table*}

\section{Experiment Costs}
\label{sec:experiment-costs}

\begin{table*}
    \centering
    \begin{tabular}{l|r|r|r|r}
        \toprule
         Dataset                  & Descriptions &  Generated tokens & Prompt tokens & Estimated cost \\
         \midrule
         YAGO 2026 (single-facts) & 4247         & 269204 & 796759 & 0.77\$ \\
         YAGO 2026 (multi-facts)  & 867          & 106502 & 255361 & 0.26\$ \\
         \bottomrule
    \end{tabular}
    \caption{Cost details for the generation of quadruple descriptions using Llama3.3-70b~\parencite{Grattafiori2024-llama3}. We made API call to the Vertex AI \texttt{llama-3.3-70b-instruct-maas} model, costing 0.72\$ per million of tokens at the time of the experiments, irrespective of whether they were generated or prompt tokens.}
    \label{tab:costs}
\end{table*}

Table~\ref{tab:costs} summarizes the cost of description generation for our dataset. In total, generating the descriptions for the YAGO 2026 TKGE dataset costed approximately $1.03$\$ for a total of 5114 examples. Generating unseen future facts using TLogic did not yield specific additional costs, and can be done on a CPU-only machine. We spent around 12 hours generating new facts using 8 CPU cores of a Linux machine.

Meanwhile, to benchmark the EDC system~\parencite{Zhang2024-edc}, we used a Google Cloud virtual machine provisioned with a single A100 40GB GPU and 256GB of storage. The experiments ran for approximately 13 hours and the hourly rate was $4.10$\$, for a total of $65.60$\$. Our requests to GPT3.5-turbo~\parencite{Brown2020-chatgpt} during the experiments amounted to approximately $0.50$\$.

Note that we only report the costs of a single run, but we had to do multiple runs during development for testing purposes.

\section{Quadruple Extraction Evaluation}
\label{sec:quadr-extr-eval}

As mentioned in the main text of the article, we adapt the WebNLG 2020 text-to-triples metrics~\parencite{Ferreira2020-webnlg} to quadruple extraction. During that adaptation, we encountered issues in the existing implementation. We think these issues warrant some discussion, since the metrics have been used to evaluate existing systems in the literature.

\subsection{Non-Optimal Alignment}

Given a reference triple $(e_s, r, e_o)$ and a candidate triple $(e_s', r', e_o')$, the \textit{exact} and \textit{partial} evaluation modes strive to compute the best possible score regardless of element type (subject, relation or object). For example, in these modes, the candidate triple $(r, e_o, e_s)$ should have a perfect score of $1$, as members of the candidate triple are simply swapped from the reference triple. However, we found that the existing implementation does not necessarily find the optimal alignment between two triples. Rather, it proceeds in a greedy manner, testing alignments in a deterministic order and returning the first partial matching one. This can lead to underestimating the score of a candidate triple. For example, we were able to construct the following adversarial example, where the \textit{exact} score should be equal to $1$:

\begin{equation}
\text{webnlg\_exact}(\text{('A B', 'B', 'C'), ('C', 'A B', 'B')}) = \frac{2}{3}
\end{equation}

We correct that issue by always returning the score of the best alignment among all the possible ones.

\subsection{Incorrect Match}

The existing implementation has a bug in some cases, where it suggests an existing match between members of the reference and candidate even though they have no substring in common. This can lead to overestimating the \textit{strict} and \textit{type} scores of a candidate triple. Similar to above, we were able to construct the following adversarial example showcasing that issue, where the \textit{strict} score should be $0$:

\begin{equation}
\text{webnlg\_strict}(\text{('A', 'B', 'C'), ('C', 'A', 'B')}) = \frac{1}{3}
\end{equation}

We correct that bug in our implementation to ensure that members can only match if they are equal (\textit{strict}) or have some substring in common (\textit{type}).

\section{Fact Descriptions Generation Prompts}
\label{sec:fact-desc-prompt}

\begin{table*}
  \centering
  \begin{tabular}{p{4cm}|p{12cm}}
    \toprule
    \textbf{Setup} & \textbf{Prompt} \\
    \midrule
    Single-fact    &
    Given the following event represented as a quadruplet of the form (subject, relation, object, timestamp):
    
    \$QUADRUPLE
    
    The following definition for the \$RELATION relation:
    
    \$RELATION\_DESCRIPTION
    
    Generate a one to three sentences description text for this event, in the style of a newspaper.
    You can add additional details, but the entirety of the information in the given quadruplet must be preserved. 
    Do NOT add any additional information or text: you must only generate the description.
    
    \{IF \$ADD\_HEADLINE
        The current date is \$CURRENT\_DATE. In addition to the date of the event, indicate the current date at the top of your text as part of a news headline. 
    \} \\
    \midrule
    Multi-facts    & Given the following events represented as quadruplets of the form (subject, relation, object, timestamp):
    
    \$QUADRUPLES
    
    and the following definitions for the relations:
    
    \$RELATION\_DEFINITIONS
    
    Generate a short paragraph describing these events, in the style of a newspaper.
    You can add additional details, but the entirety of the information in the given quadruplets must be preserved.
    Do NOT add any additional information or text: you must only generate the description.
    
    \{IF \$ADD\_HEADLINE
        The current date is \$CURRENT\_DATE. In addition to the date of the event, indicate the current date at the top of your text as part of a news headline.
    \} \\
    \bottomrule
  \end{tabular}
  \caption{Prompts used to generate unseen facts descriptions.}
  \label{tab:prompts}
\end{table*}

We present the prompts used to generate unseen facts descriptions for the single and multi-facts setups in Table~\ref{tab:prompts}. In 25\% of the cases, we add the random headlines present at the end of each prompt to increase the difficulty of the task by adding an additional timestamp in the text.

\end{document}